\documentclass[runningheads]{llncs}
\usepackage{graphicx}

\usepackage{tikz}
\usepackage{comment}
\usepackage{amsmath,amssymb} 
\usepackage{color}
\usepackage{booktabs}

\usepackage[accsupp]{axessibility}  

\usepackage[colorlinks=true,linkcolor=black,citecolor=green,urlcolor=blue,]{hyperref}
\usepackage[doipre={doi:~}]{uri}
\usepackage[numbers,sort&compress]{natbib}

\usepackage{indentfirst}
\usepackage{bbding} 
\begin{document}
\sloppy
\pagestyle{headings}
\mainmatter
\def\ECCVSubNumber{100}  

\title{1st Place Solution for CVPR2023 BURST Long Tail and Open World Challenges}
\titlerunning{CVPR-23 submission ID \ECCVSubNumber} 
\authorrunning{CVPR-23 submission ID \ECCVSubNumber} 
\author{Kaer Huang\textsuperscript{1}
}
\institute{\textsuperscript{1}Lenovo }

\maketitle

\begin{abstract}

Currently, Video Instance Segmentation (VIS) aims at segmenting and categorizing objects in videos from a closed set of training categories that contain only a few dozen of categories, lacking the ability to handle diverse objects in real-world videos. As TAO and BURST datasets release,
we have the opportunity to research VIS in long-tailed and open-world scenarios. Traditional VIS methods are evaluated
on benchmarks limited to a small number of common classes, But practical applications require trackers that go
beyond these common classes, detecting and tracking rare and even never-before-seen objects. Inspired by the latest
MOT paper for the long tail task (Tracking Every Thing in the Wild, Siyuan Li etl), we use the same idea for VIS
long-tailed and open-world setting where classification is the bottleneck. We add a mask branch on the TETer-SwinL
model for segmentation and tuned the classification loss weight to 1/100 of the original setting to ignore more effort
in optimizing classification loss. and then using the cluster method (Class Exemplar Matching) to identify classification
labels. We use BUSRT mask labels to generate ground box annotation for training and inference. for the
BURST long tail challenge, we train our model on a combination of LVISv0.5 and the COCO dataset using repeat factor sampling. First, train the detector with segmentation and CEM on LVISv0.5 + COCO dataset. And then, train the instance appearance similarity head on the TAO dataset. at last, our method (LeTracker) gets 14.9 HOTAall in the BURST test set, ranking 1st in the benchmark. for the open-world challenges, we only use 64 classes (Intersection
classes of BURST Train subset and COCO dataset, without LVIS dataset) annotations data training, and testing on
BURST test set data and get 61.4 OWTAall, ranking 1st in the benchmark. Our code will be released to facilitate future research.

\end{abstract}


\section{Introduction}

Video Instance Segmentation (VIS) aims to recognize, localize, segment, and track objects in a given video sequence.
It is a cornerstone of dynamic scene analysis and vital for many real-world applications such as autonomous driving, augmented reality, video surveillance, and short video analysis. Despite impressive efforts, Video Instance Segmentation is fundamentally constrained to segment and classify objects from a small closed set of training categories, thus limiting the capacity to generalize to large scale and diverse real-world[2]. Most VIS tasks are evaluated on benchmarks limited to a small number of common
classes. Practical applications require trackers that go beyond these common classes, detecting and tracking rare and even never-before-seen objects. As TAO and BURST datasets release, we have the opportunity to push VIS research to long-tail or open-world tasks.

\begin{figure}[ht!]
    \centering
    \includegraphics[width=0.99\linewidth]
    {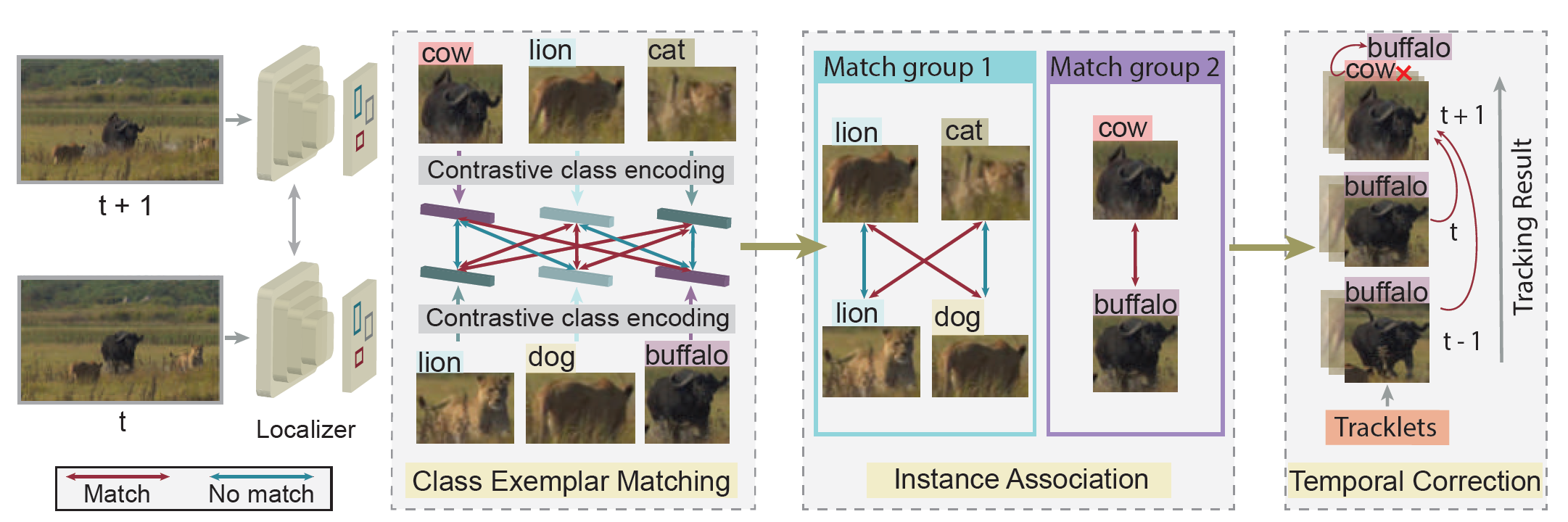}
    \caption{TETer}
    \label{fig:overview}
\end{figure}


\section{Related Work}

\textbf{Video Instance Segmentation (VIS)} is tracking all instances belonging to a fixed category set and obtaining their independent masks in every frame. As a combination of identification, tracking, and segmentation, its output consists of the corresponding predictions for different instances, each result including a category label, a
confidence score, and a binary mask sequence of this instance. Most of the early research focuses on frame-based online methods. MaskTrack R-CNN proposes an extra
tracking branch adding to the existing Mask R-CNN framework, setting a simple but strong baseline for VIS. A series of subsequent works follow this tracking-by-detection paradigm, coupled with explicit instance associations. For example, IDOL learns discriminative instance representations through comparative learning, achieving good performance
results. Although with good real-time performance, these frame-based methods lack the ability to capture long-term temporary dependency. However, this can be realized naturally in another clip-based manner, which divides frames of videos into several clips. Multi-frame modeling within the clip is performed, followed by instance
matching of the inter-clip. These offline clip-based methods achieve better performance because they capture richer spatiotemporal features. For example, VisTR uses a transformer inside the clip for parallel end-to-end sequence prediction. Despite the performance gains achieved by the above methods, they are less effective in scenarios with
congestion, occlusion, high-speed motion, and similar instances. To address this issue, MDQE and InstMove propose to use spatiotemporal priors to initialize object queries and to explicitly predict position and motion deformations respectively, providing significant improvements in the segmentation of challenging scenes.

\textbf{Long Tail Tracking} is a new research area that requires building trackers that work for rare objects, that may only contain a few examples in the training set. Contemporary MOT are designed for closed-set scenarios where all objects appear frequently in the training and testing data distributions. Hence, Dave et al proposed a new benchmark, TAO, that focuses on studying MOT in the long tail of the object category distribution. On this benchmark, AOA, QDTrack, and TET achieve impressive performance. However, those works are still limited to pre-defined object categories and thus do not scale to the diversity of real-world settings.

\textbf{Open World VIS} is a new research area that requires building trackers that can generalize to never-before-seen objects. OWTB proposed Open World Tracking to segment and track all the objects in videos. OWTB achieves state-of-the-art performance on several multi-object tracking datasets. Recent work on UVO focuses on class-agnostic object segmentation and tracking in videos, neglecting the object classification performance during evaluation. BURST[14] mainly follows the evaluation protocols of class-agnostic Multiple Object Tracking, while only the 80 training categories in MS-COCO are measured for category-wise evaluation. So methods adopted in BURST, such as STCN, have no ability to classify objects from novel categories.

\textbf{Multi Object Tracking (MOT)} is a very general algorithm and has been studied for many years. The mainstream methods follow the tracking-by-detection paradigm \cite{liang2022rethinking,lu2020retinatrack,sun2020transtrack,wu2021track}. With the development of deep learning in recent years, the performance of the detection model is improved rapidly. Currently, most of the work relies on YOLOX \cite{yan2022towards,zhang2021bytetrack}. Our method selected a stronger performance network CBNetV2 \cite{liang2021cbnetv2}  which is used to verify the potential of the detector in our hypothesis. Another important component of MOT is an association strategy. 
Popular association methods include motion-based (IoU matching, Kalman filter) \cite{bewley2016simple}, appearance-based (ReID embedding) \cite{wang2021different}, transformer-based \cite{zeng2021motr}, or the combination of them \cite{zhang2021fairmot,pang2021quasi}. Our methods remove all motion information and use only a high-performance appearance model.
\par

\textbf{Multi Object Tracking and Segmentation (MOTS)} is highly related to MOT by changing the form of boxes to fine-grained mask representation  \cite{yan2022towards}. Many MOTS methods are developed upon MOT trackers \cite{ke2021prototypical,voigtlaender2019mots,huang2023multi}. Our ideas are similar to theirs. A mask header was added on the basis of the MOT network in our MOTS solution. 

\par

\textbf{Self-Supervised Learning} has made significant progress in representation learning in recent years. Contrastive learning, one of the self-supervised learning methods such as MoCo\cite{he2020momentum}, SimCLR\cite{chen2020simple},  BYOL\cite{grill2020bootstrap},  etc, has performance that is getting closer to results of supervised learning methods in ImageNet dataset. We leveraged Momentum Contrastive Learning (MoCo-v2)\cite{he2020momentum} to train a new appearance embedding model without using tracking annotations. The technique not only meets the requirements of SSMOT and SSMOTS\cite{huang2023multi} but also improves the performance of the appearance model.

\section{Method}

\subsection{Overall Architecture}
Our method gets inspiration from the Long Tail MOT paper, which takes the cluster method to learn the classification information of the target object. As shown in Table 1, We add a mask branch to TETer for the VIS task. We also tuned the classification loss weight to 1/100 of the original setting to ignore more effort in optimizing classification loss. Then we train and infer this model on the BURST dataset. Next, we will detail explain every part of the model in the next section.

\subsection{Detection and Segmentation}

Our method has a general class-agnostic detection with segmentation, As shown in Figure 1. Detail implementation based on FasterRCNN with SwinL as the backbone. The difference with normal FasterRCNN detection is the classification branch network which changes from class num (N) to binary classes (Yes/No) which means whether it includes objects. Detailed implementation is shown in Figure 2.

\subsection{Clustering Class Exemplar Label}
Like TETer\cite{li2022tracking}, our method uses a clustering solution as a class learning module which does not depend on the detection classification branch. As shown in Figure 3.

\subsection{TETer}
As shown in Figure 4. The Main Architect of our solution
is the same as TETer except we adapt it to the VIS
task.

\section{Results}
\subsection{Dataset}
\textbf{BURST} provides 2,914 videos with pixel-precise labels for 16,089 unique object tracks (600,000 per-frame masks) spanning 482 object classes. For the Long Tail challenge which allows the use of the LVIS dataset. So we train detection and CEM module parameters by merging COCO trainset with LVISv0.5, and then, train the instance appearance similarity head on the BURST train dataset. For Open World Challenge, LVIS data is not allowed on this challenge. So we just only Use classes (64) that intersect COCO and BURST trainset.

\subsection{Metrics}
We evaluate all tasks using Higher Order Tracking Accuracy (HOTA) because it strikes a good balance between measuring frame-level detection and temporal association accuracy. For the open-world task, a slightly modified, recall-based variant of HOTA called Open World Tracking Accuracy (OWTA) is used.

\subsection{Main Results}

Our detailed results are shown in Table 2 and Table 3.
Compare with other solutions, we get both champions in
Long Tail and Open World Challenges. from the analysis
comparison results between common and uncommon classes, our solution has a big improvement with uncommon class tracking.
\begin{table*}[!ht]
    \centering
    \caption{Long Tail Result}
    \label{Long_Tail Challenge}
    \begin{tabular}{c c c c c c c c c c c }
    \hline
      Tracker   & HOTAall & DETAall & AssAall & HOTAcom & DETAcom & AssAcom & HOTAunc & DETAunc & AssAunc \\ 
    
     STCN Tracker & 4.5 & 5.4 & 4.6 & 17.1 & 19.6 & 16.7 &2.0 & 2.6 & 2.2  \\
     Box Tracker & 5.7 & 5.4 & 6.8 & 20.1 & 19.6 & 23.2 &2.9 & 2.6 & 3.6  \\
     \textbf{LeTracker(Our)}    &	\textbf{14.9} & \textbf{12.3} & \textbf{20.1} & \textbf{36.1} & \textbf{30.7} & \textbf{47.8} & \textbf{10.8} & \textbf{8.7} & \textbf{14.6}   \\

    \hline
    \end{tabular}
\end{table*}

\begin{table*}[!ht]
    \centering
    \caption{Open World Result}
    \label{Open World Challenge}
    \begin{tabular}{c c c c c c c c c c}
    \hline
      Tracker  & OWTAall & DETReall & AssAall & OWTAcom & DETRecom & AssAcom  & OWTAunc & DETReunc & AssAunc  \\ 
    \hline
     STCN Tracker2   &	57.5 & 61.6 & \textbf{54.1} & 62.9 & 71.5 & \textbf{55.7} & 23.9 & 21.0 & 28.6 \\
    
     OWTB    &	56.4 & 70.7 & 45.5 & 59.9 & 76.6 & 47.4 & 38.3 & 45.7 & 33.6  \\ 
     
     Box Tracker   &	55.9 & 61.5 & 51.1 & 61.0 & 71.4 & 52.5 & 24.6 & 21.1 & 30.0 \\ 
     \textbf{LeTracker(Our)} &	\textbf{61.4} & \textbf{74} & 51.5 & \textbf{64.3} & \textbf{78.7} & 52.9 & \textbf{47.1} & \textbf{54.6} & \textbf{42.0} \\

    \hline
    \end{tabular}
\end{table*}

\section{Conclusition}
Traditional VIS methods are evaluated on benchmarks
limited to a small number of common classes and do not
have long tail issues, But practical applications require trackers that go beyond these common classes, detecting and tracking rare and even never-before-seen objects. Inspired by the latest MOT paper for the long tail task (Tracking Every Thing in the Wild, Siyuan Li et al), we use the same idea for VIS long-tailed and open-world setting where classification is the bottleneck. We add a mask branch on the TETer-SwinL model for segmentation and tuned the classification loss weight to 1/100 of the original setting to ignore more effort in optimizing classification loss. and then using the cluster method (Class Exemplar Matching) to identify classification labels. We use BUSRT mask labels to generate ground box annotation for training and inference.
for the BURST long tail challenge, we train our model on a combination of LVISv0.5 and the COCO dataset using repeat factor sampling. First, train the detector with segmentation and CEM on LVISv0.5 + COCO dataset. And then, train the instance appearance similarity head on the TAO train dataset. at last, our method (LeTracker) gets 14.9 HOTAall in the BURST test set, ranking 1st in the benchmark. for the open-world challenges, we only use 64 classes (Intersection classes of BURST Train subset and COCO dataset, without LVIS dataset) annotations data training and testing on BURST test set data and get 61.4 OWTAall, ranking 1st in the benchmark.

\renewcommand{\bibname}{References}
\bibliographystyle{splncs04}
\bibliography{egbib}
\end{document}